# Bioinformatics and Classical Literary Study

**Pramit Chaudhuri[1*], Joseph P. Dexter[2*]**

1 University of Texas at Austin, USA

2 Harvard University, USA

*Corresponding authors: Pramit Chaudhuri (pramit.chaudhuri@austin.utexas.edu), Joseph P. Dexter (jdexter@fas.harvard.edu)

**Abstract**
This paper describes the Quantitative Criticism Lab, a collaborative initiative between classicists, quantitative biologists, and computer scientists to apply ideas and methods drawn from the sciences to the study of literature. A core goal of the project is the use of computational biology, natural language processing, and machine learning techniques to investigate authorial style, intertextuality, and related phenomena of literary significance. As a case study in our approach, here we review the use of sequence alignment, a common technique in genomics and computational linguistics, to detect intertextuality in Latin literature. Sequence alignment is distinguished by its ability to find inexact verbal similarities, which makes it ideal for identifying phonetic echoes in large corpora of Latin texts. Although especially suited to Latin, sequence alignment in principle can be extended to many other languages.

**keywords**
intertextuality; sequence alignment; Latin; literature; computational biology; natural language processing

## 0 INTRODUCTION

The Quantitative Criticism Lab (QCL; www.qcrit.org), directed by the co-authors, grew out of shared interests in classical intertextuality dating back to a seminar taught in 2009. The idea for a collaborative and interdisciplinary study of intertextuality was first discussed by us in January 2014, inspired locally by Dexter's work in computational biology and globally by the rise of the Digital Humanities and "big data" approaches to cultural study. We have since published articles on a range of topics in both humanities and science journals, and QCL has received support from the National Endowment for the Humanities, the Andrew W. Mellon Foundation, the American Council of Learned Societies, and the Neukom Institute for Computational Science. QCL maintains a physical lab space at the University of Texas at Austin, but the project team includes scholars from a number of different institutions, in addition to undergraduate research assistants and high school students and teachers.

The impetus for bridging the disciplines of biology and literary study is both cultural and methodological. The cultural motive derives from the familiar analogies often drawn between literature and living organisms, whether the use of trees or woods as a metaphor for literary works and traditions, or the imaginative casting of poems and books as living embodiments or descendants of their author. Such analogies have a long pedigree and have taken many different forms: one famous instance occurs at the end of Ovid's *Metamorphoses*, when the poet conceives of his own life as inextricably entwined with the life of his epic, a work which itself thematizes the mutually implicated transformation of bodies and words. Although the comparisons across these two domains are of countless types and have distinctive local functions in the particular works, collectively they invite critics to scrutinize the basic analogy: in what way can literature and living organisms profitably be linked?

The recent explosion of work in the study of biological and, in particular, evolutionary processes has exerted an effect on literary study too: from cognitive humanists viewing literature as a revealing expression of evolutionary psychology (Winter 2012 issue of *Critical Inquiry*) to researchers using phylogenetic analysis to generate stemmata of manuscripts (Howe and Windram, 2011). More recently, QCL has used stylometry and machine learning to obtain nuanced portraits of the cultural evolution of literature, focusing especially on two richly imitative strands of the Latin literary tradition - pseudo-Senecan and Senecan influenced tragedy, and citations of earlier fragmentary historians in Livy's monumental history of Rome (Dexter *et al.,* 2017). Although diverse in their humanistic aims, these applications of biology tend to mirror one of two principal approaches that have long been characteristic of literary study, historicism and formalism. The former regards literature as distinctive evidence for mental and social phenomena; the latter, showing greater affinity with the text-life analogies above, regards literature primarily in terms of its shapes, sounds, and resemblances. As has often been pointed out, however, this distinction is an artificial one, and criticism in practice frequently - even necessarily - traffics between both modes. It remains to be seen whether the productive dialogue between historicism and formalism can have any meaningful equivalent in the manifold evolutionary approaches to literature, a question that can perhaps only be answered once the field has had considerably longer to reflect on developments that have been as rapid as they are far-reaching.

With that broad brush sketch of the state of the discipline in mind, the goals of this project presentation are rather more modest and specific. The main object of the presentation is to describe a tool we have developed for intertextuality detection that exploits a technique from computational biology and linguistics known as sequence alignment. Following the binary characterization of literary criticism above, our enhancement of intertextual search methods has more in common with formalist than historicist criticism. However, this consideration should not be taken as a commitment to formalism *tout court*, since newly discovered intertextuality can and should be put in the service of historicist interpretations.

Although the tool itself represents only one part of the project's work, it exemplifies our broader intellectual motivation and aims. After a brief contextualization of the approach in relation to other computational studies of culture, we explain the methodology of sequence alignment, and discuss case studies in its initial application to intertextuality in Latin epic poetry and tragedy.

## I From Big Data to the Single Datum

### 1.1 Humanistic Applications of "Big Data"
A glance at any popular science publication will remind one that biology is in an age of "-omics". Genomics, of course, most famously - but also transcriptomics, proteomics, metabolomics, and so on. The rise of "-omics" has been driven by technologies (such as next-generation sequencing of DNA and RNA) that allow many measurements to be made quickly and in parallel, and has relied on computational techniques that enable researchers to pinpoint features of interest in vast data sets. Although computational biology operates at a larger scale than classical philology - for instance, the human genome contains approximately 3.2 billion bases, equivalent in length to about 9,000 copies of the *Aeneid* - a central premise of our work is that bioinformatics has much to offer digital Classics. Our aim in this section is to offer some very brief reflections on the types of humanistic questions researchers outside of Classics have addressed using high-throughput methods, and then to introduce our own humanistic appropriation of one particular method.

An influential example of such an interdisciplinary approach to culture is a 2011 paper in *Science* by Erez Lieberman Aiden, Jean-Baptiste Michel, and colleagues, which reported a quantitative analysis of the whole of Google Books, roughly 4% of all text ever printed (Michel *et al.*, 2011). Their method, which they dubbed "culturomics", is to trace the frequency of "n-grams" (sequences of "n" characters or words) across time. Applications of the method have included spotting trends in linguistic change over time, pinpointing moments of concentrated interest in some topic or other, and identifying periods of censorship in cultural history. Follow-up n-gram analyses of more targeted corpora, such as British periodicals, have yielded additional insights (Lansdall-Welfare *et al.,* 2017).

Big data approaches such as culturomics have had great success in illuminating large-scale linguistic trends. A thornier issue has been the integration of such methods with more traditional modes of literary criticism (Dexter *et al.,* 2017), which often requires the identification of points of interest within texts that can stimulate further critical inquiry. Here bioinformatic techniques can be of particular value. A fundamental computational task in genomics is the identification of similar ("homologous") gene sequences. Biological homology is never exact, however, and searching for identically matching sequences, although computationally easy, is therefore not useful. The study of literary intertextuality confronts an analogous problem: exact repetition of phrases constitutes only one (especially overt) class of intertextuality, and an ideal computational method should be sensitive to less obvious resemblances. In the next section, we describe how sequence alignment, now universally employed for finding gene homologues, can be applied to this problem.

**1.2 Sequence Alignment and Fīlum**
Sequence alignment is a method to compare two strings of arbitrary characters on a character-by-character basis (Needleman and Wunsch, 1970, Wagner and Fischer, 1974, Smith and Waterman, 1981). "Alignment" means to position the strings side-by-side so that they match at as many positions as possible. One metric for indicating the similarity or dissimilarity between pairs of aligned strings is known as the "edit distance". This score equals the minimum number of single-character changes (additions, deletions, or substitutions) required to convert string A into string B. The ubiquitous bioinformatics tool BLAST (Basic Local Alignment Search Tool) uses a variant of sequence alignment to identify homologues, in conjunction with various computational heuristics required for efficient searching of entire genomes (Altschul *et al.,* 1990). When sequence alignment is applied to natural language texts, a simple edit distance calculation can quantify similarity between words or phrases. For example, the edit distance between A = "*arma virumque*" and B = "*arma virique*" is two because two changes (substitution of "i" for "u" and deletion of "m") are required to turn A into B. In this case, string A is the famous opening of Vergil's *Aeneid*, while string B is a similar phrase used by a later Roman epic poet, Silius Italicus, to allude back to the *Aeneid* (*Punica* 6.6; Landrey, 2014).

In literary study, the diversity of intertextual connections is a major challenge for any method of detection, whether traditional or computational: within the full range of intertextuality, quotations (typically exact matches) are in the minority and are more straightforward to identify than resemblances that can range from close (e.g., phrases distinguished only by morphological variation, as in the example above) to distant (e.g., phrases similar in sound but only partially similar in diction, as we will encounter below). Given the overlap between *arma virumque* and *arma virique*, it is fairly intuitive to see how sequence alignment can provide a useful and efficient means of discovering non-exact intertextual resemblances. With this aim in mind, we have developed an intertextual search tool based on sequence alignment. The tool, called Fīlum ("string" in Latin), takes as inputs a query phrase and target corpus and then generates a list of phrases in the corpus whose edit distance from the query is below a user-specified cutoff. Users may also specify whether to search for phrases of a fixed word order or to conduct an order-free search, in which the constituent words are aligned individually. Order-free searches can be applied to adjacent words or to words separated by intervening words (the number of which is designated throughout by the term "interval"). The need for such latitude is especially important in the context of Latin verse, since word order is highly flexible and poets often reuse or adapt phrases without maintaining the same sequence as in the source text. As noted above, literary texts are much shorter than genomes, so Fīlum computes exact edit distances and does not rely on any heuristics. Although sequence alignment has been used for textual comparison before, it has been employed at the level of word rather than character (Olsen *et al.*, 2010). Using character-by-character alignment, however, enables the detection of phrases that are only partly similar, thereby offering a significant advantage in identifying non-exact parallels of potential literary interest. A web-based implementation of Fīlum with an intuitive interface is available on the QCL website (https://qcrit.org/filum).

### 1.3 Case Study 1: Classical Latin Epic

Classicists have benefited from the availability of multiple computational tools to aid intertextual search. Two well-known examples include Diogenes, a word search tool that has long been an integral component of the philologist's toolkit, and the core tool designed by the Tesserae Project, an innovative approach to find repeated two-word phrases between pairs of texts (Coffee *et al.*, 2013). We envision sequence alignment as a useful complement to Diogenes and Tesserae, which can effectively identify words in common across one or more texts. The facility of both tools to capture morphological variants results in an improved signal-to-noise ratio in comparison to sequence alignment's intentionally more indiscriminate method. The great success of Tesserae, for instance, was demonstrated in a 2012 study that reported a systematic tracing of the reuse of Vergil's *Aeneid* in Lucan's epic *Pharsalia* (Coffee *et al.*, 2012).

One particular strength of sequence alignment is well illustrated by an example discovered in the course of testing our tool, which likewise is drawn from the *Aeneid* and *Pharsalia*. We searched for parallels to the query *commune nefas* ("collective crime"), a key phrase in the opening of the *Pharsalia* (1.6). Searching a corpus of five epics and ten tragedies at a maximum edit distance of three and with fixed word order yields 11 results, of which the two most interesting are *commune nefas* (i.e., the identical phrase) in the *Thyestes* (139), a tragedy by Lucan's uncle Seneca, and *immane nefas* ("immense crime") in the *Aeneid* (6.624). While Lucan may have drawn inspiration from both sources, it is the Vergilian phrase that highlights a benefit of sequence alignment. The words *commune* and *immane* are similar in sound alone, but attention to the context of each phrase suggests that Lucan is deliberately alluding to Vergil, as well as Seneca, in his diction. Attempting to search for the common word *nefas* in the same corpus generates 256 instances, which illustrates the advantage of using sequence alignment to identify a manageable but still interesting set of possible intertexts. Finally, we note that the Vergilian intertext is not amenable to discovery using the Tesserae method, since the phrases have in common only one, not two, words. One strength of sequence alignment is thus the identification of parallels sharing common sounds, whether morphological endings or more general phonetic resemblances, which helps to fill out the profile of creative variation across texts.

The project's first major publication appeared in a 2015 special issue of the Classics journal *Dictynna* devoted to Latin epic poetry of the Flavian period (AD 69-96), an important focal point for our project and for the broader study of Roman literature (Chaudhuri *et al*., 2015). The survival of a cluster of texts from the same genre and time period affords literary critics a rare opportunity to see intertextuality at work among closely affiliated epic poems and their principal canonical models. The *Dictynna* article introduces the method of sequence alignment for intertextual search, taking as a case study the intertextuality between Silius Italicus' *Punica* (a Flavian epic about the Hannibalic or Second Punic War) and its two main models, the *Aeneid* and Livy's prose history of Rome. The article compares sequence alignment with other computational methods of intertextual search and examines a series of computationally identified parallels that have not been discussed in commentaries or other scholarship.

The *Dictynna* article offers another good example of an intertext based in part on sound. Since we discuss the passage at length in the article, we confine ourselves here to the salient details and a new addition. The Silian phrase in question is *Acheronta videbo* ("I will see Acheron [a river in hell]", *Punica* 2.367), which in context expresses the Carthaginian speaker Gestar's willingness to die rather than being enslaved to Rome. The words are an unmistakable echo of Vergil's Juno, who famously declares *Acheronta movebo* ("I will move Acheron [i.e., the powers of hell]", *Aeneid* 7.312), at the beginning of the second half of the *Aeneid*. In brief, the significance of the allusion rests on two factors. First, as patron goddess of Carthage, Juno possesses a deep hostility towards its eventual destroyer, Rome, an attitude that extends even to her use of nether powers against the Trojans, whose descendants will eventually found Rome. Juno's Vergilian expression then finds its way into a similar utterance by Gestar, one of Silius' Carthaginian characters and a staunch opponent of Rome. Second, whereas Juno as a goddess is able to command hell, Gestar is a mere human for whom hell means little more than death: the change of verb is thus instrumental in characterizing the speaker and his relatively limited agency.

In an extension to our prior study we ran a new search for Silius' phrase *Acheronta videbo* against the *Aeneid* using a maximum edit distance of six and with fixed word order. The results reveal that in addition to the main echo of Juno, Silius may have combined Vergilian reminiscences, drawing the change of verb from another related phrase, *Simoenta videbo* ("I will see Simois [a river at Troy]") at *Aeneid* 5.634. This secondary echo comes from a speech by Iris, an agent of Juno, who seeks to burn the Trojan fleet. While Silius probably did not intend any direct reference to the passage in *Aeneid* 5, it is plausible that he was aware of the connection between the two Vergilian passages and reflected that in his own diction. The new result further demonstrates the utility of sequence alignment, since Silius' two Vergilian intertexts have no words in common with each other, but each phrase shares a single word with Silius' phrase, augmented by the phonetic similarity of the other word (*-nta*, *-ebo*).

## 1.4 Case Study 2: Neo-Latin Tragedy

The value of computational searches is particularly great in the case of less well-known corpora that do not have the extensive critical resources available for most classical literature. The neo-Latin works of early modernity, for instance, represent a large and important body of material that remains understudied and lacks the detailed intertextual characterizations scholars have produced for Vergil and other classical writers. Computational searches for intertextuality provide a rapid and straightforward means to characterize a neo-Latin text's connections to ancient or other post-classical material. In this way, techniques such as sequence alignment can not only aid the research of neo-Latin specialists, but also open the door to classicists wishing to explore new areas of Latin literature.

To illustrate the potential usefulness of Fīlum for the study of non-canonical, neo-Latin, material, we searched a sample of the *Progne*, a 15th century tragedy by Gregorio Correr. The sample consisted of the play's 67-line first speech, which we profiled using queries drawn from Seneca's *Thyestes*. Correr explicitly cites the *Thyestes* as a model for his composition, a relationship corroborated by substantial intertextual connections. Many of these imitations are so close as to require little more than exact word searches, for example, *Sisyphi ... lapis* ("the stone of Sisyphus") occurring at both *Progne* 12 and *Thyestes* 6. One result of our searches, however, illustrates the particular strength of sequence alignment already suggested by the Silius and Lucan examples discussed above. We searched for the phrase *dignum facinus* ("worthy crime", *Thyestes* 271) using a maximum edit distance of four and a maximum interval of two. Of the four results, one seemed of plausible literary significance: *dirum facinus* ("terrible crime") at *Progne* 36, at an edit distance of two and interval of one. Not only does Correr echo the sound and phrasing of the *Thyestes*, he also deploys the words in a similar context. Just as the speaker in the *Thyestes*, Atreus, contemplates the extremity of his revenge against his brother, so too the speaker in the *Progne*, Diomedes, considers the extremity of the vengeance about to be enacted in the drama. Despite this general resemblance, however, the change in epithet signals a difference in the two speakers' attitudes: whereas Diomedes appears to regret his own misdeeds and to be distressed by the impending crimes of his descendants and their resulting infamy, Atreus by contrast sees wrongdoing - especially unprecedented wrongdoing - as a worthy goal.

The *Progne* also reworks numerous other Latin texts, including Senecan tragedies besides the *Thyestes*. For this reason, we also searched for Correr's phrase *dirum facinus* using the same parameters (maximum edit distance of four, maximum interval of two) against a corpus of all 10 dramas associated with Seneca, including two composed by imitators shortly after Seneca's death. This yielded 81 results in total, including (as expected) the phrase from the *Thyestes*, *dignum facinus*. One other result, however, stood out: the same phrase as in the *Progne*, *dirum facinus*, appears at line 931 of Seneca's *Medea* (at an edit distance of zero, but with the words in reverse order, thus demonstrating the utility of the order-free search function). Although low edit distance is a marker of similarity rather than of significance, additional qualitative analysis suggests the parallel's literary relevance for the *Progne*. The phrase is used by the protagonist Medea to describe the act of killing her own children as she debates whether to perform the deed. This act is the same one that Diomedes envisages at the opening of the *Progne*, since the eponymous mother Progne will, like Medea, kill her son to avenge herself on her husband. Correr thus makes a single phrase do double work: on the one hand, it suggests a general resemblance to, yet appreciable difference from, his main Senecan model, the *Thyestes*; on the other hand, it echoes an important secondary model, the *Medea*, that is in some ways a more exact paradigm for the particular telos of this play.

The two searches involving the *Progne* thus show the capacity of Fīlum for uncovering intertexts of multiple types, both exact matches and partially similar phrases. In this case, the searches revealed a combinatorial allusion whereby Correr was able to make literarily meaningful use of two source passages simultaneously. As with Lucan's *commune nefas* and Silius' *Acheronta videbo*, sequence alignment is able to capture a distinctly poetic aspect of Correr's intertexuality, in which phrases that share only one word in common can have a greater affinity because of phonetic or rhythmic similarity (*dignum facinus ~ dirum facinus*). This phenomenon warrants further study and could potentially have a bearing on the future development of Fīlum, which might be modified to take greater account of phonetic matching (rather than operating on the basis of simple edit distance).

**Conclusion**

The contrasting methodologies underlying the various tools used for intertextual search allow for the capture of a broad range of comparanda and, as a consequence, a similarly broad range of analyses. That combination of multiple forms of intertextual search, analysis, and interpretation puts the scholar of Latin literature in an unusually good position to answer both microscopic and macroscopic questions about a text and its relationships to other texts. This computationally- and critically-informed approach represents, in our view, a plausible model of what most Latin literary scholarship of the future will look like. The application of sequence alignment to intertextuality also exemplifies the dual contribution that biology can make to literary criticism - as a rich source for advanced computational methods that work equally well for genomes and natural language texts, and as a framework for understanding the vast networks uncovered by intertextual study.

**Acknowledgements**

Many people have contributed to QCL since its inception in 2014. We would like to acknowledge several participants in particular. Two primary collaborators, Tathagata Dasgupta and Nilesh Tripuraneni, are principally responsible for providing expertise on machine learning and its application to natural language texts. The third primary collaborator, Ariane Schwartz, is principally responsible for providing expertise on classical literature and its reception. Edward Gan and Ajay Kannan helped to develop an initial implementation of Fīlum, and Jorge Bonilla Lopez contributed to the analysis of Silius Italicus' *Punica* referenced above. QCL has been supported by funding from the Office of the Provost at Dartmouth College, a Neukom Institute for Computational Science CompX Faculty Grant, and a National Endowment for the Humanities Digital Humanities Start-Up Grant (HD-248410-16). J.P.D. has been supported by a National Science Foundation Graduate Research Fellowship (DGE1144152), and P.C. by an American Council of Learned Societies Digital Innovation Fellowship and an Andrew W. Mellon Foundation New Directions Fellowship.